\documentclass{article}
\usepackage{spconf,amsmath,graphicx,mathtools}
\usepackage[utf8]{inputenc}

\usepackage{float}
\restylefloat{table}

\title{ON MODELING ASR WORD CONFIDENCE}
\name{Woojay Jeon, Maxwell Jordan, and Mahesh Krishnamoorthy}
\address{Apple Inc.\\ One Apple Park Way, Cupertino, California\\ \small \tt{\{woojay,maxwell\_jordan,maheshk\}@apple.com}}
\newcommand\blfootnote[1]{%
  \begingroup
  \renewcommand\thefootnote{}\footnote{#1}%
  \addtocounter{footnote}{-1}%
  \endgroup
}

\begin{document}
\ninept
\abovedisplayskip=4pt
\belowdisplayskip=4pt
\abovecaptionskip=3pt
\belowcaptionskip=2pt
\maketitle
\begin{abstract}
We present a new method for computing ASR word confidences that effectively mitigates the effect of ASR errors for diverse downstream applications, improves the word error rate of the 1-best result, and allows better comparison of scores across different models.
We propose 1) a new method for modeling word confidence using a Heterogeneous Word Confusion Network (HWCN) that addresses some key flaws in conventional Word Confusion Networks, and 2) a new score calibration method for facilitating direct comparison of scores from different models.
Using a bidirectional lattice recurrent neural network to compute the confidence scores of each word in the HWCN, we show that the word sequence with the best overall confidence is more accurate than the default 1-best result of the recognizer, and that the calibration method can substantially improve the reliability of recognizer combination.
\end{abstract}

\section{Introduction}
\label{sec:intro}

Automatic speech recognition (ASR) systems often output a confidence score \cite{wessel-2001} for each word in the recognized results.
The confidence score is an estimate of how likely the word is to be correct, and can have diverse applications in tasks that consume ASR output.

We are particularly interested in ``data-driven'' confidence models \cite{li-2019} that are trained on recognition examples to learn systematic ``mistakes'' made by the speech recognizer and actively ``correct'' them.
A major limitation in such confidence modeling methods in the literature is that they only look at Equal Error Rate (EER) \cite{wessel-2001} or Normalized Cross Entropy (NCE) \cite{siu-1997} results and do not investigate their impact on speech recognizer accuracy in terms of Word Error Rate (WER).
Some past studies \cite{wessel-2000}\cite{evermann-2000} have tried to improve the WER using confidence scores derived from the word lattice by purely mathematical methods, but to our knowledge no recent work in the literature on statistical confidence models has reported WER.

If a confidence score is true to its conceptual definition (the probability that a given word is correct), then it is natural to expect that the word sequence with the highest combined confidence should also be at least as accurate as the default 1-best result (obtained via the MAP decision rule).
One reason that this may not easily hold true is that word confidence models, by design, often try to force all recognition hypotheses into a fixed segmentation in the form of a Word Confusion Network (WCN) \cite{mangu-2000}\cite{tur-2003}\cite{li-2019}.
While the original motivation of WCNs was to obtain a recognition result that is more consistent with the WER criterion, we argue that it often unnaturally decouples words from their linguistic or acoustic context and makes accurate model training difficult by introducing erroneous paths that are difficult to resolve.

Instead, we propose the use of a ``Heterogeneous'' Word Confusion Network (HWCN) for confidence modeling that can be interpreted as a representation of multiple WCNs for a given utterance.
Although the HWCN's structure itself is known, our interpretation of HWCNs and our application of them to data-driven confidence modeling is novel.
We train a bidirectional lattice recurrent neural network \cite{ladhak-2016}\cite{li-2019} to obtain confidence values for every arc in the HWCN. 
Obtaining the sequence with the best confidence from this network results in better WER than the default 1-best.
In addition, we recognize the need to be able to directly compare confidence scores between different independently-trained confidence models, and propose a non-parametric score calibration method that improves the recognition accuracy when combining the results of different recognizers via direct comparison of confidence scores.

\section{Confidence modeling using Heterogeneous Word Confusion Networks}

\subsection{Defining confidence and addressing key flaws in WCNs}
\label{subsec:wcn}

Word Confusion Networks (WCNs) \cite{mangu-2000} were originally proposed with the motivation of transforming speech recognition hypotheses into a simplified form where the Levenshtein distance can be approximated by a linear comparison of posterior probabilities, thereby optimizing the results for WER instead of Sentence Error Rate (SER).

Subsequent works on word confidence modeling \cite{li-2019} have based their models on WCNs because of their linear nature, which allows easy identification and comparison of competing words.
However, WCNs are fundamentally flawed in that they force all hypothesized word sequences to share the same time segmentation, even in cases where the segmentation is clearly different.

\begin{figure}[t]
  \centering
  \includegraphics[width=2.8in]{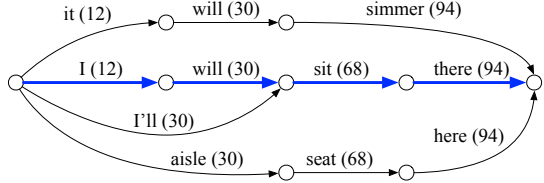}
  \vspace{-.12in}
  \caption{Word hypothesis lattice where each arc is labeled with a word and a number indicating the point in time (in frames) where the word ends. The best path ``I will sit there'' is marked in blue.}
  \label{fig:word_hyp_lattice}
\end{figure}

\begin{figure}[t]
  \centering
  \vspace{-.15in}
  \includegraphics[width=2.8in]{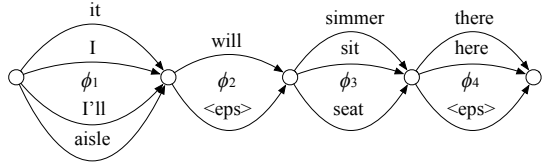}
  \vspace{-.12in}
  \caption{Word confusion network (WCN) corresponding to Fig. \ref{fig:word_hyp_lattice}, where the time segments (in frame ranges) are assigned based on the best path: $\phi_1=(0, 12), \phi_2=(12, 30), \phi_3=(30, 68), \phi_4=(68, 94)$}
  \label{fig:wcn}
  \vspace{-.20in}
\end{figure}

Consider the (contrived) word hypothesis lattice in Fig.\ref{fig:word_hyp_lattice} with 1-best sequence ``I will sit there.''
A corresponding WCN is in Fig.\ref{fig:wcn}, obtained by aligning all possible sequences with the 1-best sequence.

Let us define the confidence for a word $w$ in a time segment $\phi _j$ given acoustic features $X$ as
\begin{equation}\label{eq:confidence}
{\textrm{Word \ Confidence \ }} \buildrel \Delta \over = P\left( {\left. {{w},{\phi _j}} \right|X} \right),
\end{equation}
where $\phi _j$ is a tuple of start and end times of the $j$'th slot, drawn from a finite set of time segments, and the set $\Phi  = \left\{ {{\phi _1}, \cdots ,{\phi _n}} \right\}$ is a full description of the time segmentation of the WCN in Fig. \ref{fig:wcn}.

For a sequence of $n$ words $w_1, \cdots, w_n$ in the time segments $\phi _1, \cdots, \phi _n$, consider a random variable $N_i$ denoting the number of correct words in each slot $i$ with time segment $\phi _i$ containing $w_i$. The expectation of $N_i$ is directly the confidence probability:
\begin{equation}
E\left[ {{N_i}} \right] = P\left( {\left. {{w_i},{\phi _i}} \right|X} \right).
\end{equation}
Let the random variable $N$ denote the total number of correct words in the sequence.
In a manner similar to \cite{mangu-2000}, we approximate the WER as the ratio between the expected number of incorrect words and the total number of words. Since $E\left[ N \right] = \sum_{i = 1}^n {E\left[ {{N_i}} \right]} $, 
\begin{equation}\label{eq:wer}
\textrm{WER} \approx \frac{{n - E\left[ N \right]}}{n} = 1 - \frac{1}{n}\sum\limits_{j = 1}^n {P\left( {\left. {{w_j},{\phi _j}} \right|X} \right)}.
\end{equation}
Hence, the WER can be minimized by finding, for each slot $\phi_j$, the word with the highest ${P\left( {\left. {{w_j},{\phi _j}} \right|X} \right)}$. 

\begin{figure}[]
  \centering
  \includegraphics[width=2.1in]{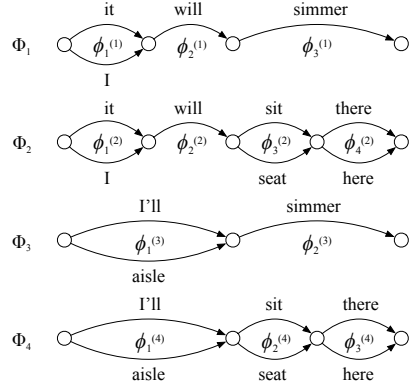}
  \vspace{-.15in}
  \caption{Four WCNs extracted from Fig.\ref{fig:word_hyp_lattice} that represent all possible word sequences without requiring forced slotting or $<$eps$>$ insertion. Each WCN has a time segmentation $\Phi _n$, where each time segment is represented by $\phi^{(n)} _m$, where $\phi^{(1)}_1 \hspace{-2pt} = \hspace{-2pt} \phi^{(2)}_1 \hspace{-2pt} = (0, 12), \phi^{(1)}_2 \hspace{-2pt} = \hspace{-2pt} \phi^{(2)}_2 \hspace{-2pt} = (12, 30), \phi^{(1)}_3 \hspace{-2pt} = \hspace{-2pt} \phi^{(3)}_2 \hspace{-2pt} = (30, 94), \phi^{(3)}_1 \hspace{-2pt} = \hspace{-2pt} \phi^{(4)}_1 \hspace{-2pt} = (0, 30), \phi^{(2)}_3 \hspace{-2pt} = \hspace{-2pt} \phi^{(4)}_2 \hspace{-2pt} = (30, 68)$ and $\phi^{(2)}_4 \hspace{-2pt} = \hspace{-2pt} \phi^{(4)}_3 \hspace{-2pt} =(68, 94)$}
  \label{fig:separate_wcns}
  \vspace{-.20in}
\end{figure}

However, in Fig.\ref{fig:wcn}, all sequences are required to follow the same time segmentation as the 1-best result, so ``I'll'' and ``aisle'' have been unnaturally forced into the same slot as ``it'' and ``I'', even though they actually occupy a much greater length of time extending into the second slot.
Also, in order to be able to encode the hypotheses ``I'll sit there'' and ``aisle seat here'', an epsilon ``skip'' label ``$<$eps$>$'' had to be added to the second slot.
Such heuristics add unnecessary ambiguity to the data that make it difficult to model.
Furthermore, confidences like $P\left({\left.{aisle,{\phi _1}} \right|X} \right)$ are actually 0, since ``aisle'' does not even fit into $\phi _1$, so no meaningful score can be assigned to ``aisle'' even though it is a legitimate hypothesis.

\subsection{Mitigating flaws in a WCN by using multiple WCNs}
\label{subsec:multiple_wcns}

To address the aforementioned problems, let us consider deriving \emph{multiple} WCNs from the lattice.
Fig.\ref{fig:separate_wcns} shows four different WCNs that represent all possible sequences in Fig.\ref{fig:word_hyp_lattice}, but without any unnaturally-forced slotting or epsilon insertion.

Each WCN has a unique segmentation ${\Phi _k} = \left\{ {\phi _1^{\left( k \right)}, \cdots ,\phi _{{n_k}}^{\left( k \right)}} \right\}$ $(k=1,\cdots,4)$ with length $n_k$, but note that a lot of the time segments are shared, i.e., $\phi^{(1)}_1 = \phi^{(2)}_1$, $\phi^{(1)}_2 = \phi^{(2)}_2$, $\phi^{(1)}_3 = \phi^{(3)}_2$, $\phi^{(3)}_1 = \phi^{(4)}_1$, $\phi^{(2)}_3 = \phi^{(4)}_2$ and $\phi^{(2)}_4 = \phi^{(4)}_3$. For every $\Phi _k$ the WER for a sequence of words $w_1, \cdots, w_{n_k}$ is
\begin{equation}\label{eq:phi_wer}
\textrm{WER}\left( {{w_1}, \cdots ,{w_{{n_k}}},{\Phi _k}} \right) = 1 - \frac{1}{{{n_k}}}\sum\limits_{j = 1}^{{n_k}} {P\left( {\left. {{w_j},\phi _j^{\left( k \right)}} \right|X} \right)}.
\end{equation}
To find the best word sequence, we simply look for the sequence of words across all four segmentations with the lowest WER:
\begin{equation}\label{eq:best_seq}
{{\hat w}_1}, \cdots ,{{\hat w}_{{n_l}}},{{\hat \Phi }_l} = \arg \mathop {\max }\limits_{{w_1}, \cdots ,{w_{{n_k}}},{\Phi _k}} \textrm{WER} \left( {{w_1}, \cdots ,{w_{{n_k}}},{\Phi _k}} \right).
\end{equation}
The WER in \eqref{eq:phi_wer} and \eqref{eq:best_seq} is more sensible than the WER in \eqref{eq:wer} because it no longer contains invalid probabilities like $P(aisle, \phi_1 | X)$ nor extraneous probabilities like $P( < \hspace{-4pt} eps \hspace{-4pt} >, \phi_2 | X)$. At the same time, it still retains the basic motivation behind WCNs of approximating Levenshtein distances with linear comparisons. 

\subsection{The Heterogeneous Word Confusion Network}
\label{subsec:hwcn}

We now propose using a ``Heterogeneous'' Word Confusion Network (HWCN). The HWCN is derived from the word hypothesis lattice by 1) merging all nodes with similar times (e.g. within a tolerance of 100ms), and then 2) merging all competing arcs (arcs sharing the same start and end node) with the same word identity.

\begin{figure}[]
  \centering
  \includegraphics[width=2.8in]{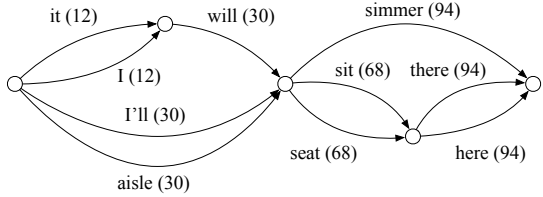}
  \vspace{-.15in}
  \caption{The Heterogeneous Word Confusion Network corresponding to Fig. \ref{fig:word_hyp_lattice}. Nodes with the same times have been merged, and competing arcs with the same word (``will'') have been merged.}
  \vspace{-.1in}
  \label{fig:hwcn}
\end{figure}

The HWCN corresponding to the lattice in Fig.\ref{fig:word_hyp_lattice} is shown in Fig.\ref{fig:hwcn}.
Since the destination nodes of ``it'' and ``I'' have the same time (12), the two nodes are merged into one.
The destination nodes of the two ``will'' arcs are also merged, making the two arcs competing arcs. 
Since they have the same word, the two arcs are merged.

\begin{figure}[]
  \centering
  \includegraphics[width=2.0in]{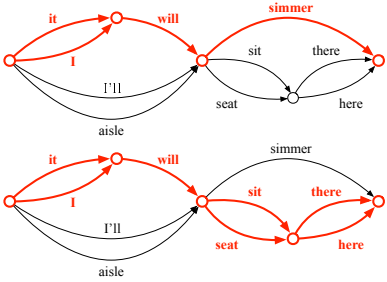}
  \vspace{-.15in}
  \caption{Illustration of how the HWCN is actually an encoding of the individual WCNs in Fig.\ref{fig:separate_wcns}. (Top) Shown in red, the first WCN (with segmentation $\Phi _1$) of Fig.\ref{fig:separate_wcns} is encoded in the upper part of the HWCN. (Bottom) Similarly, the second WCN (with segmentation $\Phi _2$) is encoded.}
  \vspace{-.15in}
  \label{fig:hwcn2}
\end{figure}

This sort of partially-merged network has been used in previous systems in different contexts \cite{wessel-2000}, but not with data-driven confidence models.
A key interpretation we contribute in this work is that \emph{the HWCN is in fact a representation of the four separate WCNs} in Fig.\ref{fig:separate_wcns}. 
Fig.\ref{fig:hwcn2} shows how the first and second WCNs of Fig.\ref{fig:separate_wcns}, with segmentations $\Phi_1$ and $\Phi_2$, respectively, are encoded inside the HWCN. 
The third and fourth WCNs can be easily identified in a similar manner.
The shared time segments (e.g. $\phi^{(1)}_1 = \phi^{(2)}_1$ in Fig. \ref{fig:separate_wcns}) are also fully represented in the HWCN.

Note that there are some extraneous paths in the HWCN that are not present in the lattice. 
For example, word sequences like ``I will simmer'' or ``aisle sit there'' can occur. 
On the other hand, if what the speaker actually said was ``I will sit here'', which is not a possible path in the lattice, the HWCN has a chance to correct eggregious errors in the recognizer's language model to provide the correct transcription.

Now, if we train a word confidence model to compute the scores in Eq.\eqref{eq:confidence} for every arc in the HWCN, the WER in Eq.\eqref{eq:best_seq} can be minimized by finding the sequence in the HWCN with the highest mean word confidence per Eq.\eqref{eq:phi_wer} via dynamic programming.

When merging arcs, such as the two ``will'' arcs in Fig.\ref{fig:word_hyp_lattice}, we must define how their scores will be merged, as these scores will be used as features for the confidence model. 
Assume $n$ arcs $e_1, \cdots, e_n$ to be merged, all with the same word, start time, and end time, and consuming the same acoustic features $X$. Each $e_i$ starts at node $v_i$ and ends at $v'_i$, and has an arc posterior probability ${P\left( {\left. {{e_i}} \right|X} \right)}$, acoustic likelihood ${p\left( {\left. {X} \right|{e_i}} \right)}$, and transitional (language \& pronunciation model) probability $P\left( {\left. {{e_i}} \right|{v_i}} \right)$. 
We want to merge the start nodes into one node $v$, the end nodes into one node $v'$, and the arcs into one arc $e$. The problem is to compute $P\left( {\left. e \right|X} \right)$, $p\left( {\left. {X} \right|e} \right)$, and $P\left( {\left. e \right|v} \right)$.

If $e$ conceptually represents the union of the $n$ arcs, the posterior of the merged arc $e$ is the sum of the posteriors of the individual arcs.
This is because we only merge competing arcs (after node merging), and there is no way to traverse two or more competing arcs simultaneously (e.g. the two ``will'' arcs in Fig.\ref{fig:word_hyp_lattice}), so the traversal of such arcs are always disjoint events.
The acoustic scores of the original arcs should be very similar since their words are the same (but may have different pronunciations) and occur at the same time, so we can approximate the acoustic score of the merged arc as the mean of the individual acoustic scores. Hence,
\begin{equation}\label{eq:merged_post}
P\left( {\left. e \right|X} \right) = \sum\limits_{i = 1}^n {P\left( {\left. {{e_i}} \right|X} \right)} ,\,\,p\left( {\left. X \right|e} \right) \approx \frac{1}{n}\sum\limits_{i = 1}^n {p\left( {\left. X \right|{e_i}} \right)}.
\end{equation}
The transitional score of the merged arc can be written as
\begin{equation}
P\left( {\left. e \right|v} \right) = \sum\limits_{i = 1}^n {P\left( {\left. {e,{v_i}} \right|v} \right)}  = \sum\limits_{i = 1}^n {P\left( {\left. e \right|{v_i},v} \right)P\left( {\left. {{v_i}} \right|v} \right)}.
\end{equation}
It is easy to see that the first term in the summation is $P\left( {\left. e \right|{v_i},v} \right) = P\left( {\left. e \right|{v_i}} \right) = P\left( {\left. {{e_i}} \right|{v_i}} \right)$.
As for the second term, if $v$ represents the union of the $n$ nodes, we have
\begin{equation}
P\left( {\left. {{v_i}} \right|v} \right) = \frac{{P\left( {{v_i}} \right)}}{{\sum\limits_{j = 1}^n {P\left( {{v_j}} \right)} }}
\end{equation}
where $P(v_i)$ is the prior transitional probability of node $v_i$ that can be obtained by a lattice-forward algorithm on the transitional scores in the lattice.

When labeling the arcs of the HWCN as correct(1) or incorrect(0) for model training, we align the 1-best sequence with the reference sequence.
Then, for any arc in the 1-best that has one or more competing arcs, we label each competing arc as 1 if its word matches the corresponding reference word or 0 if it does not.
All other arcs in the HWCN are labeled 0.
If there is no match between the 1-best and the reference sequence, all arcs are labeled 0.

\section{Confidence score calibration}
\label{sec:calibration}

While our abstract definition of confidence is Eq.\eqref{eq:confidence}, the confidence score from model $i$ is actually an estimate according to $\lambda _i$, i.e., $\widehat P\left( {\left. {W,\phi } \right|X;{\lambda _i}} \right)$ where $\lambda _i$ are the parameters of the ASR and the confidence model trained on HWCNs.
In general, it is problematic to directly compare the probability estimates computed from two different independently-trained statistical models. 
This poses a problem in a personal assistant when a given downstream natural language processor (that uses the ASR's confidence values) is expected to work consistently regardless of updates to the ASR.

To investigate this problem, we consider a naive classifier combination method which, given $n$ different classifiers, simply chooses the result with the highest reported confidence score (such as for personalized ASR \cite{paulik-2018}).
Our goal is to calibrate the confidence scores so that such a direct comparison would work.
It is easy to prove that when the confidence scores are the true probabilities of the words being correct, the system will always be as least as accurate overall as the best individual classifier.
Suppose each classifier $k$ outputs results with confidence $C_k$ and hence has expected accuracy $E[C_k]$ (since $C_k$ itself is the probability of being correct). 
The combined classifier has confidence $C=\max \{ C_1, \cdots, C_n \}$ with expected accuracy $E\left[ C \right] = E\left[ {\max \left\{ {{C_1}, \cdots ,{C_n}} \right\}} \right] \geq E\left[ {{C_k}} \right]$ for all $k$, so it is at least as good as the best classifier.
When the confidence scores are not the true probabilities, however, this guarantee no longer holds, and such a naive system combination may actually degrade results.

In this work, we propose a data-driven calibration method that is based on distributions of the training data but do not require heuristic consideration of histogram bin boundaries \cite{zadrozny-2001} nor make assumptions of monotonicity in the transformation \cite{zadrozny-2002}.

Given a confidence score of $y \in \mathcal{R}$, we consider the probability $P(E_c | y)$ where $E_c$ is the event that the result is correct.
We also define $E_w$ as the event that the result is wrong, and write
\begin{equation} \label{eq:true_prob}
P\left( {\left. {{E_c}} \right|y} \right) = \frac{{p\left( {\left. y \right|{E_c}} \right)P\left( {{E_c}} \right)}}{{p\left( {\left. y \right|{E_c}} \right)P\left( {{E_c}} \right) + p\left( {\left. y \right|{E_w}} \right)P\left( {{E_w}} \right)}}.
\end{equation}
The priors $P(E_c)$ and $P(E_w)$ can be estimated using the counts $N_c$ and $N_w$ of correctly- and incorrectly-recognized words, respectively, over the training data, i.e., $P\left( {{E_c}} \right) \approx \frac{{{N_c}}}{{{N_c} + {N_w}}}$ and $P\left( {{E_w}} \right) \approx \frac{{{N_w}}}{{{N_c} + {N_w}}}$.

We use the fact that the probability distribution function is the derivative of the cumulative distribution function (CDF):
\begin{equation} \label{eq:derivative}
p\left( {\left. y \right|{E_c}} \right) = \frac{d}{{dy}}p\left( {\left. {Y \le y} \right|{E_c}} \right).
\end{equation}
One immediately recognizes that the CDF is in fact \emph{the miss probability of the detector at threshold} $y$:
\begin{equation} \label{eq:pm}
p\left( {\left. {Y \le y} \right|{E_c}} \right) = {P_M}\left( y \right).
\end{equation}

$P_M$ can be empirically estimated by counting the number of positive samples in the training data that have scores less than $y$:
\begin{equation}
{P_M}\left( y \right) \approx \frac{1}{{{N_c}}}\sum\limits_{i \in {I_c}}^{} {u\left( {y - {y_i}} \right)}.
\end{equation}
where $I_c$ is the set of (indices of) positive training samples, $y_i$ is the confidence score of the $i$'th training sample, and $u(x)$ is a step function with value 1 when $x>0$ and 0 when $x \le 0$.
In order to be able to take the derivative, we approximate $u(x)$ as a sigmoid function controlled by a scale factor $L$:
\begin{equation}
u\left( x \right) \approx \frac{1}{{1 + {e^{ - xL}}}}.
\end{equation}

This lets us solve Eq.\eqref{eq:derivative} to obtain
\begin{equation}\label{eq:pc}
p\left( {\left. y \right|{E_c}} \right) = \frac{1}{{{N_c}}}\sum\limits_{i \in {I_c}}^{} {\frac{{L{e^{\left( {{y_i} - y} \right)L}}}}{{{{\left\{ {1 + {e^{\left( {{y_i} - y} \right)L}}} \right\}}^2}}}}.
\end{equation}

Likewise, we can see that $p\left( {\left. y \right|{E_w}} \right)$ is the negative derivative of the \emph{false alarm probability of the detector at threshold} $y$:
\begin{equation}\label{eq:pfa}
p\left( {\left. y \right|{E_w}} \right) = \frac{d}{{dy}}p\left( {\left. {Y \le y} \right|{E_w}} \right) = - \frac{d}{{dy}}{P_{FA}(y)},
\end{equation}
which leads to
\begin{equation}\label{eq:pw}
p\left( {\left. y \right|{E_w}} \right) = \frac{1}{{{N_w}}}\sum\limits_{j \in {I_w}}^{} {\frac{{L{e^{\left( {{y_j} - y} \right)L}}}}{{{{\left\{ {1 + {e^{\left( {{y_j} - y} \right)L}}} \right\}}^2}}}},
\end{equation}
where $I_w$ is the set of (indices of) negative training samples.
Applying Eqs. \eqref{eq:pc} and \eqref{eq:pw} to Eq.\eqref{eq:true_prob} now gives us a closed-form solution for transforming the confidence score $y$ to a calibrated probability $P\left( {\left. {{E_c}} \right|y} \right)$, where the only manually-tuned parameter is $L$.
In our experiments, we found $L=1.8$ to work the best on the development data.

\section{Experiment}

We took 9 different U.S.-English speech recognizers used at different times in the past for the Apple personal assistant Siri, each with its own vocabulary, acoustic model, and language model, and trained lattice-RNN confidence models \cite{li-2019} on the labeled HWCNs of 83,860 training utterances. 
For every speech model set, we trained a range of confidence models -- each with a single hidden layer with 20 to 40 nodes and arc state vectors with 80 to 200 dimensions -- and took the model with the best EER on the development data of 38,127 utterances. 
The optimization criterion was the mean cross entropy error over the training data. 
For each arc, the features included 25 GloVe Twitter \cite{pennington-2014} word embedding features, a binary silence indicator, the number of phones in the word, the transitional score, the acoustic score, the arc posterior, the number of frames consumed by the arc, and a binary feature indicating whether the arc is in the 1-best path or not. 
The evaluation data was 38,110 utterances.

To evaluate detection accuracy, we compare the confidence scores with the arc posteriors obtained from lattice forward-backward computation and merged via Eq.\eqref{eq:merged_post}.
Tab.\ref{tab:detection} shows the EER and NCE values, computed using the posteriors and confidences on the labeled HWCNs on the evaluation data.

Next, we measure the WER when using the path with the maximum mean word confidence (which minimizes the WER per Eq.\eqref{eq:best_seq} and Sec.\ref{subsec:hwcn}). 
Tab.\ref{tab:wer} shows that the WER decreases for every recognizer in this case, compared to using the default 1-best result. 
The WER decrease is marginal in some cases, but there is a decrease in all 9 recognizers, implying that the effect is statistically significant.

Finally, we assess the impact of the score calibration in Sec.\ref{sec:calibration}.
There are $2^9-10=502$ possible combinations of recognizers from the nine shown in Tab.\ref{tab:wer}.
Combination of $n$ recognizers is done by obtaining $n$ results (via best mean confidence search on the HWCN of every recognizer) and choosing the result with the highest mean word confidence.
\begin{table}[H]
\centering
\begin{tabular}{ l | c | c | c }
 Method    & EER (\%) & NCE\footnotemark \\
\hline
  Arc Posterior from Forward-Backward    & 4.23 & 0.621 \\
  Proposed Confidence     & 3.42 & 0.868 \\
\end{tabular}
\caption[]{EER and NCE for the arc posterior from lattice-forward-backward, and the proposed confidence measure on evaluation data. \vspace{-.1in}}
\label{tab:detection}
\end{table}

\footnotetext{Erroneously-swapped values have been corrected after ICASSP 2020}

\begin{table}[H]
\vspace{-.18in}
\centering
\setlength\tabcolsep{4pt} 
\begin{tabular}{ l |c| c | c | c | c | c | c | c | c}
   & 1 & 2 & 3 & 4 & 5 & 6 & 7 & 8 & 9 \\
\hline
 B & 12.26 & 6.93 & 4.95 & 4.88 & 4.80 & 4.88 & 10.92 & 6.81 & 6.46 \\
 P & 12.00 & 6.84 & 4.90 & 4.85 & 4.76 & 4.84 & 10.86 & 6.75 & 6.43 \\
\end{tabular}
\caption[]{WER (\%) of nine recognizers on the evaluation data. 
The Baseline (B) uses the default 1-best result obtained from the MAP decision rule, while the Proposed (P) uses the word sequence with the maximum mean word confidence. \vspace{-.1in}}
\label{tab:wer}
\end{table}

\begin{table}[H]
\vspace{-.18in}
\centering
\begin{tabular}{ l | c | c }
   & Raw & Calibrated \\
\hline
 No. of times better & 110 (21.9\%) & 502 (100\%) \\
 No. of times worse & 392 (78.1\%)  &  0 (0\%) \\
\end{tabular}
\caption[]{Impact of score calibration on system combination.
Out of a total 502 experiments, we counted the number of times the combined system did better or worse than the best individual system when using the raw confidence and the calibrated confidence. \vspace{-.1in}}
\label{tab:comb1}
\end{table}

\begin{table}[H]
\vspace{-.18in}
\centering
\begin{tabular}{ l | c | c | c }
  Recognizers & Best Indiv. & Raw conf. & Calib. conf. \\
  Combined    & WER(\%)     & WER(\%)  & WER(\%)    \\
\hline 
 2, 7             & 6.84 (no.2) & 7.24 & 6.80 \\
 3, 4, 5, 6, 9    & 4.76 (no.5) & 4.39 & 4.63 \\
 1, 5, 6, 8, 9    & 4.76 (no.5) & 5.33 & 4.61 \\
\end{tabular}
\caption[]{Sample recognizer combination results from the experiment in Tab.\ref{tab:comb1}, showing the best individual WER among the recognizers combined, the WER when combining using raw confidences, and the WER when combining using calibrated confidences. \vspace{-.1in}}
\label{tab:comb2}
\end{table}
We performed all combinations, and counted the number of times the WER of the combined system was better than  the best individual WER of the recognizers used in each combination.

Tab.\ref{tab:comb1} shows that when using the ``raw'' confidence scores from the RNN model, in most cases (78.1\%) the combined recognizer had higher WER.
When the calibrated scores proposed in Sec.\ref{sec:calibration} are used, however, the combined system beat the best recognizer in all trials.
Tab.\ref{tab:comb2} shows some example combination results. 
Anecdotally, we found the WER from raw confidences tend to have higher variance than the WER from calibrated confidences. 
The raw scores sometimes give very accurate results, but Tab.\ref{tab:comb1} shows the calibrated scores give improvements much more consistently.

\section{Conclusion and future work}
We have proposed a method for modeling word confidence using Heterogeneous Word Confusion Networks and showed that they have better detection accuracy than lattice arc posteriors as well as improving the WER compared to the 1-best result from the MAP decision rule. We have also proposed a method for calibrating the confidence scores so that scores from different models can be better compared, and demonstrated the efficacy of the method using system combination experiments.

Future work could address one shortcoming of the proposed method in that there is no normalization of the confidence scores to ensure that $\textstyle\sum_{w, \phi} {P\left( {\left. {w,\phi } \right|X} \right)}  = 1$.
\blfootnote{Thanks to Rogier van Dalen, Steve Young, and Melvyn Hunt for helpful comments.}

\bibliographystyle{IEEEbib}
\bibliography{strings,refs}

\end{document}